\documentclass[lettersize,journal]{IEEEtran}

\usepackage[T1]{fontenc}
\usepackage{graphicx}
\usepackage[caption=false]{subfig}
\usepackage{mathrsfs,amsmath,amsfonts,amssymb,amsthm,amstext,amscd,amsxtra,amsopn}
\usepackage{diagbox}
\usepackage{lettrine}
\usepackage{cite}
\usepackage{eucal}
\usepackage{soul}
\providecommand{\keywords}[1]{\textbf{\textit{Index terms---}} #1}
\usepackage{bm}
\usepackage{multirow}
\usepackage{esint}
\usepackage{empheq}
\usepackage{comment}
\usepackage{boxhandler}
\usepackage{bm}
\usepackage{algorithm}
\usepackage[noend]{algpseudocode}
\usepackage{etoolbox}
\apptocmd{\sloppy}{\hbadness 10000\relax}{}{}
\usepackage{leftidx}
\usepackage{makecell}
\usepackage{stackrel}
\usepackage{xcolor}
\usepackage{hyphenat}
\usepackage{nicefrac}
\usepackage[normalem]{ulem}
\usepackage[hyphens]{url}
\usepackage[hidelinks]{hyperref}
\hypersetup{breaklinks=true}
\algnewcommand{\IfThenElse}[3]{
  \State \algorithmicif\ #1\ \algorithmicthen\ #2\ \algorithmicelse\ #3}
\algnewcommand{\IfThen}[3]{
  \State \algorithmicif\ #1\ \algorithmicthen\ #2}

\makeatletter
\def\BState{\State\hskip-\ALG@thistlm}

\makeatother

\algnewcommand{\Initialize}[1]{%
  \State \textbf{Initialize:}
  \Statex \hspace*{\algorithmicindent}\parbox[t]{.8\linewidth}{\raggedright #1}
}

\let\oldref\ref
\renewcommand{\ref}[1]{(\oldref{#1})}

\makeatletter
\DeclareRobustCommand{\pder}[1]{%
  \@ifnextchar\bgroup{\@pder{#1}}{\@pder{}{#1}}}
\newcommand{\@pder}[2]{\frac{\partial#1}{\partial#2}}
\makeatother

\begin{document}
\title{PIFON-EPT: MR-Based Electrical Property Tomography Using Physics-Informed Fourier Networks}

\author{Xinling Yu, José E. C. Serrallés, \IEEEmembership{Member, IEEE}, Ilias I. Giannakopoulos, Ziyue Liu, Luca Daniel, \IEEEmembership{Fellow, IEEE}, Riccardo Lattanzi, \IEEEmembership{Senior Member, IEEE}, Zheng Zhang, \IEEEmembership{Member, IEEE}

\thanks{This work was supported in part by research grants from the National Science Foundation (NSF 2107321) and the National Institute of Health (NIH 2R01 EB024536) in the United States.
\textit{(Corresponding author: Xinling Yu)}}
\thanks{Xinling Yu and Zheng Zhang are with the Department of Electrical and Computer Engineering, University of California, Santa Barbara, CA 93106 USA (email: xyu644@ucsb.edu).}
\thanks{José E. C. Serrallés and Luca Daniel are with the Research Laboratory of Electronics, Department of Electrical Engineering and Computer Science, Massachusetts Institute of Technology, Cambridge, MA 02139 USA.}
\thanks{Ilias I. Giannakopoulos is with the Bernard and Irene Schwartz Center for Biomedical Imaging, Department of Radiology, New York University Grossman School of Medicine, NY 10016 USA.}
\thanks{Riccardo Lattanzi is with the Center for Advanced Imaging Innovation and Research (CA$\text{I}^2$R), and with the Bernard and Irene Schwartz Center for Biomedical Imaging, Department of Radiology, New York University Grossman School of Medicine, NY 10016 USA.}
\thanks{Ziyue Liu is with the Department of Computer Science, University of California, Santa Barbara, CA 93106 USA.}}
\bstctlcite{IEEEexample:BSTcontrol}
\maketitle

\begin{abstract}

We propose Physics-Informed Fourier Networks for Electrical Properties (EP) Tomography (PIFON-EPT), a novel deep learning-based method for EP reconstruction using noisy and/or incomplete magnetic resonance (MR) measurements. Our approach leverages the Helmholtz equation to constrain two networks, responsible for the denoising and completion of the transmit fields, and the estimation of the object’s EP, respectively. We embed a random Fourier features mapping into our networks to enable efficient learning of high-frequency details encoded in the transmit fields. We demonstrated the efficacy of PIFON-EPT through several simulated experiments at 3 and 7 tesla (T) MR imaging, and showed that our method can reconstruct physically consistent EP and transmit fields. Specifically, when only $20\%$ of the noisy measured fields were used as inputs, PIFON-EPT reconstructed the EP of a phantom with $\leq 5\%$ error, and denoised and completed the measurements with $\leq 1\%$ error. Additionally, we adapted PIFON-EPT to solve the generalized Helmholtz equation that accounts for gradients of EP between inhomogeneities. This yielded improved results at interfaces between different materials without explicit knowledge of boundary conditions. PIFON-EPT is the first method that can simultaneously reconstruct EP and transmit fields from incomplete noisy MR measurements, providing new opportunities for EPT research.

\end{abstract}

\keywords{\textbf{Electrical Property Mapping, Fourier Features Mapping, Magnetic Resonance Imaging, Physics Informed Neural Networks.}}


\section{Introduction} \label{Introduction}
\IEEEPARstart{E}{lectrical} properties (EP), namely relative permittivity and electric conductivity, determine the interactions between electromagnetic waves and biological tissue \cite{collins2011calculation, hand2008modelling}. EP have the potential to be employed as biomarkers for pathologies such as cerebral ischemia \cite{holder1992detection, fallert1993myocardial} and cancer \cite{tha2014noninvasive, tha2018noninvasive, balidemaj2015feasibility, shin2015initial}. EP could also be used to improve the effectiveness of existing therapeutic modalities such as radiofrequency hyperthermia \cite{rossmann2014review, balidemaj2016hyperthermia,de2016bio}.

Several EP tomography (EPT) methods have been proposed that are based on MR measurements, such as the magnetic transmit ($B_{1}^{+}$) or receive ($B_{1}^{-}$) field maps \cite{haacke1991extraction, wen2003noninvasive, katscher2009determination, voigt2011quantitative,hafalir2014convection,zhang2010imaging, sodickson2012local,serralles2019noninvasive, leijsen20183,  liu2015gradient, gurler2017gradient}. These techniques can be classified based on the form of Maxwell's equations (differential or integral) they use to fit the MR measurements. Differential methods, such as the Helmholtz EPT (H-EPT) \cite{katscher2009determination} or the Convection-Reaction EPT (CR-EPT) \cite{hafalir2014convection}, require the calculations of spatial derivatives of noisy measured $B_1^{+}$ maps, which lead to errors and artifacts in the reconstructions \cite{mandija2018error}. On the other hand, integral equation-based methods \cite{leijsen20183, serralles2019noninvasive} are robust to noise, but require computationally expensive iterative optimizations that rely on an accurate model of the transmit coils \cite{giannakopoulos2019global, giannakopoulos2020magnetic} and fine-tuned regularization parameters.

Recently, data-driven deep learning-based methods have been introduced for EP reconstruction \cite{mandija2019opening, hampe2020investigating, gavazzi2020deep, giannakopoulosusage} to mitigate the noise amplifications and high computational cost of standard methods. These methods treat MR measurements and EP distributions as 2D images or 3D volumes, and train regression convolution neural networks as surrogate EP reconstruction models from simulated training data. These supervised learning-based techniques perform well in simulation, but they are not reliable in vivo due to the necessarily limited number of different cases included in the training data. To improve the generalization to in-vivo data, hybrid techniques that embed deep learning into conventional EP mapping methods were proposed \cite{leijsen2022combining, inda2022physics1}. These hybrid methods use neural networks to generate initial guesses of EP for iterative reconstruction schemes \cite{leijsen2022combining}, or diffusion and convection coefficients for the convection-reaction equation \cite{inda2022physics1}. While these approaches improve generalization, several electromagnetic simulations are still required to generate training data, which can be very expensive and time-consuming, thus there is only a limited amount of available datasets. A recent hybrid technique directly reconstructs conductivity from input transceive phases \cite{inda2022physics2}. In such a method, a neural network is trained to represent the input transceive phase map, where the gradients of the phase are computed by automatic differentiation \cite{baydin2018automatic} and then used to solve the phase-only convection-reaction EPT. The reconstructed conductivity is compared with ground-truth values at the boundary, as a regularization for the neural network that represents the phase. Since this method retains the physics of EPT, it does not require a comprehensive set of electromagnetic simulations. However, learning a single neural network that can simultaneously represent the ground-truth phase and provide accurate gradient approximations directly from noisy measured phase maps is challenging, which is shown by the fact that they yielded highly inaccurate EP reconstructions in most cases. 

Following our preliminary study \cite{yu2022mr}, here we propose the Physics-Informed Fourier Networks (PIFONs) Electrical Properties Tomography (PIFON-EPT) framework, which leverages recent developments on physics-informed deep learning \cite{raissi2019physics,karniadakis2021physics,qi2020two,qi2023electromagnetic}, and Fourier features mapping \cite{tancik2020fourier} to learn both the EP distribution and the $B_{1}^{+}$ field globally from noisy and/or incomplete $\tilde{B}_{1}^{+}$ measurements. The proposed framework can efficiently de-noise the $\tilde{B}_{1}^{+}$ measurements. Once trained, PIFONs can accurately predict the EP and $B_{1}^{+}$ field at any location within the PDE domain, enhancing high-resolution imaging capabilities. In contrast to integral equation-based methods \cite{leijsen20183, serralles2019noninvasive}, which necessitate repeated simulations of forward equations, PIFONs tackle the inverse problem directly. This approach has the computational cost equivalent to solving a single forward equation. Differently from other supervised learning-based EPT methods \cite{mandija2019opening, hampe2020investigating, gavazzi2020deep, giannakopoulosusage}, our proposed PIFON-EPT technique can reconstruct EP directly, without being trained on known $B_{1}^{+}$ and EP distribution pairs. Compared with recent physics-aware hybrid EPT methods \cite{inda2022physics1, inda2022physics2} in which EP are still solved numerically from convection-reaction equation with boundary condition, our method represents EP as a neural network constrained by the Helmholtz equations and does not require any prior EP information.


The rest of the paper is organized as follows: In Section \ref{Technical Background}, we provide a brief overview of standard EPT methods. In Section \ref{Methods}, we describe the proposed novel PIFON-EPT framework. In Section \ref{Results}, we demonstrate the effectiveness of our PIFON-EPT with four representative numerical experiments. Further discussion is provided in Section \ref{Discussion}, whereas Section \ref{Conclusion} summarizes the main points of this work.

\section{Technical Background} \label{Technical Background}
\subsection{Fundamental Helmholtz Equations in MRI}
The relation between the magnetic field ($\mathbf{B}$) and the EP of a medium can be described by the Helmholtz equation:
\begin{equation}
\nabla^2 \mathbf{B} + k_{0}^2 \varepsilon_c \mathbf{B}+\nabla \varepsilon_c \times \frac{\nabla \times \mathbf{B}}{\varepsilon_c} = 0,
\label{eq1}
\end{equation}
where $k_{0}$ is the wave number in vacuum and
\begin{equation}
\varepsilon_{c} = \varepsilon_{r} - \frac{i\sigma}{\omega\varepsilon_{0}},
\label{eq_ep}
\end{equation}
is the relative complex permittivity. Here, $\varepsilon_{r}$ is the relative permittivity and $\sigma$ is the electric conductivity, $i$ denotes the imaginary unit, $\omega$ denotes the angular frequency, and $\varepsilon_{0}$ denotes the vacuum permittivity. Since the full transmit $\mathbf{B}_1$ cannot be measured in an MRI scanner, but only its positively rotating component $B_{1}^{+} = (B_{x} + iB_{y})/2$, we can re-write equation \eqref{eq1} with the help of Gauss' law ($\nabla \cdot \mathbf{B}=0$) as:
\begin{equation}
\small
\begin{aligned}
\nabla^2 B_1^{+} +  k_{0}^2 \varepsilon_c B_1^{+} &=
\left(\frac{\partial B_1^{+}}{\partial x}-i \frac{\partial B_1^{+}}{\partial y}+\frac{1}{2} \frac{\partial B_z}{\partial z}\right)\left(g_x+i g_y\right) \\
&+\left(\frac{\partial B_1^{+}}{\partial z}-\frac{1}{2} \frac{\partial B_z}{\partial x}-i \frac{1}{2} \frac{\partial B_z}{\partial y}\right) g_z.
\label{generalized Heq}
\end{aligned}
\end{equation}
Here, $\mathbf{g}:=\left(g_x, g_y, g_z\right):=\nabla \ln \varepsilon_c$. If we assume a smooth distribution of the EP, their gradient $\mathbf{g}$ can be ignored, and equation \eqref{generalized Heq} becomes the \textit{homogeneous Helmholtz equation}:
\begin{equation}
\nabla^2 B_1^{+} + k_{0}^2 \varepsilon_c B_1^{+} = 0.
\label{Heq}
\end{equation}

\subsection{Standard Differential EPT Methods}
\label{subsec:EPT}

One can solve equations \eqref{generalized Heq} and \eqref{Heq} for the EP, starting from measured $B_1^{+}$ maps. There are several methods based on such approach (here is a non-exhaustive list \cite{katscher2009determination, voigt2011quantitative,hafalir2014convection,zhang2010imaging, sodickson2012local,serralles2019noninvasive, leijsen20183,  liu2015gradient}). Next, we provide a brief overview of two popular ones: the Helmholtz EPT \cite{katscher2009determination} and the Convection-Reaction EPT \cite{hafalir2014convection}. Both techniques require the knowledge of absolute phase of $B_1^{+}$, which, for birdcage coils, can be estimated with the transceive
assumption \cite{katscher2009determination}. Open-source software implementations of these methods can be found in EPTlib \cite{arduino2021eptlib}.

\subsubsection{Helmholtz EPT}
Assuming a homogeneous distribution of the EP and access to measured complex $\tilde{B}_1^{+}$ maps, one can directly invert the homogeneous Helmholtz equation \eqref{Heq} to estimate the EP:
\begin{equation}
\varepsilon_c  = -\frac{\nabla^2 \tilde{B}_1^{+}}{k_{0}^2 \tilde{B}_1^{+}}.
\label{HEPT}
\end{equation}
The second-order spatial derivatives of the measured $\tilde{B}_1^{+}$ can be computed via finite difference approaches. If the measured fields are noisy, smoothing filters such as the $2^{\text{nd}}$ order Savitzky-Golay filter \cite{savitzky1964smoothing} can be applied to improve the numerical derivatives.

\subsubsection{Convection-Reaction EPT}
High-field MRI scanners (< 7 T) utilize birdcage-based body coils \cite{hayes1985efficient} for transmission. In these cases, the $B_{z}$ component of the coil's magnetic field can be assumed negligible near the mid-plane of the scanner bore. As a result, the generalized Helmholtz equation \eqref{generalized Heq} can be simplified as:
\begin{equation}
\begin{aligned}
\nabla^2 \tilde{B}_1^{+} +  k_{0}^2 \varepsilon_c \tilde{B}_1^{+} &=
\left(\frac{\partial \tilde{B}_1^{+}}{\partial x}-i \frac{\partial \tilde{B}_1^{+}}{\partial y} \right)\left(g_x+i g_y\right) \\
&+\frac{\partial \tilde{B}_1^{+}}{\partial z} \cdot g_z.
\label{generalized Heq without Bz}
\end{aligned}
\end{equation}
If we let $\gamma = 1/\varepsilon_{c}$, equation \eqref{generalized Heq without Bz} can be rewritten as the convection-reaction equation with a zero diffusion term with respect to $\gamma$ \cite{hafalir2014convection}:
\begin{equation}
\small
\begin{aligned}
\nabla^2 \tilde{B}_1^{+} \cdot \gamma +  k_{0}^2  \tilde{B}_1^{+} &=
-\left(\frac{\partial \tilde{B}_1^{+}}{\partial x}-i \frac{\partial \tilde{B}_1^{+}}{\partial y} \right)\left(\frac{\partial \gamma}{\partial x}+i \frac{\partial \gamma}{\partial y}\right) \\
&-\frac{\partial \tilde{B}_1^{+}}{\partial z} \cdot \frac{\partial \gamma}{\partial z}.
\label{Cr eq}
\end{aligned}
\end{equation}
By imposing appropriate boundary conditions (for example, the value of $\gamma$ at the boundary of the domain), the convection-reaction equation \eqref{Cr eq} can be solved with a mesh-based finite difference scheme for $\gamma$. As for Helmholts EPT, also in this case the gradients of the measured $\tilde{B}_1^{+}$ can be estimated using the Savitzky-Golay filter \cite{savitzky1964smoothing}. Since at MRI frequencies below 3 T, the absolute phase of $B_{1}^{+}$ is almost independent from the permittivity \cite{wen2003noninvasive}, it is possible to perform conductivity-only reconstructions using only the absolute phase of $\tilde{B}_1^{+}$ \cite{gurler2017gradient}. It is also possible to include an artificial diffusion term to the convection-reaction equation to stabilize and improve the reconstruction results \cite{li2017mr}. 

\section{Methods} \label{Methods}
\begin{figure*}[ht]
    \centering
    \includegraphics[scale=0.6]{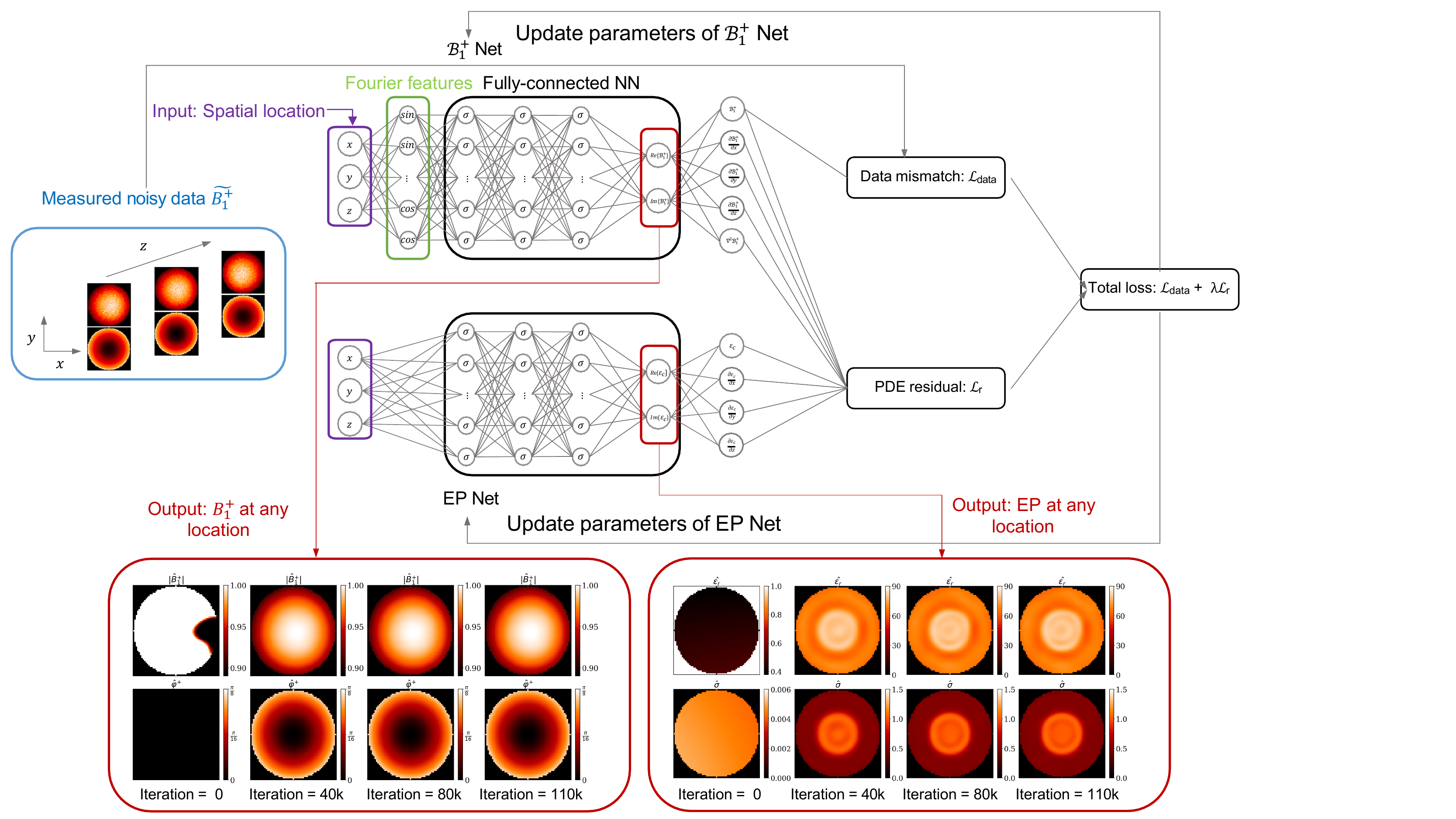}
    \caption{PIFON-EPT workflow. Two separate fully-connected neural networks $B_{1}^{+}$ Net ($\mathcal{B}_{1}^{+}(\bm{r};\bm{\theta_{1}})$) and EP Net ($\mathcal{E}_{c}(\bm{r};\bm{\theta_{2}})$) are defined to take spatial coordinates $\bm{r}=(x,y,z)$ as inputs and output the corresponding $B_{1}^{+}$ field and the EP distributions, respectively at the same $\bm{r}$ locations. The $B_{1}^{+}$ Net and EP Net are trained jointly by minimizing a composite loss function that aims to fit the measured $\tilde{B}_{1}^{+}$ data (blue dotted box) while also penalizing the PDE residual. Once trained, the resulting physics-informed $B_{1}^{+}$ Net and EP Net can be used to obtain physically consistent predictions of $B_{1}^{+}$ and EP at any arbitrary 3D location. A representative axial cut of the outputs of the neural networks obtained at different iterations during training is shown at the bottom (red dotted box).}
    \label{fig:PIFON}
\vspace{-10pt}
\end{figure*}

Our proposed PIFON-EPT is a deep learning-based framework for robust EP estimation using noisy and/or incomplete complex-valued MR measurements. Note that since in MRI we do not have direct access to the absolute phase of $B_1^{+}$, we can rely on symmetry assumptions to estimate the complex-valued field in actual experiments. Specifically, at 1.5 and 3 tesla (T), when RF birdcage coils are used for transmission and reception in quadrature, the $B_1^{+}$ and $B_1^{-}$ phases are approximately equal \cite{wen2003noninvasive,katscher2009determination}. Therefore, since the transceive phase is measurable \cite{brown2014magnetic}, we can approximate the absolute phase of $B_1^{+}$ as half the transceive phase. The goal of PIFON-EPT is to learn the EP distributions globally that best describe the complex-valued $B_{1}^{+}$ at any spatial location $(x, y, z)$, given $\{(\bm{r_{i}}, \tilde{B}_{1}^{+}(\bm{r_{i}}))\}_{i=1}^{N}$ only for a limited $N$ locations $\bm{r_i} = (x_{i},y_{i},z_{i})$. The workflow of PIFON-EPT is summarized in Fig. \ref{fig:PIFON}.

\subsection{PIFON-EPT workflow}
Traditional EPT methods based on finite difference approximation of derivatives of $B_{1}^{+}$ \eqref{HEPT}, \eqref{Cr eq} can lead to noise amplifications in the reconstructed EP distributions. To prevent this, we seek to solve an optimization problem constrained by the measured data and physical laws using physics-informed deep learning \cite{raissi2019physics}. We denote the Helmholtz equation that describes the physical laws that must be satisfied by $B_{1}^{+}$ in the general form on a d-dimension domain $\Omega \in \mathbb{R}^{d}$:
\begin{equation}
\mathcal{N}_{\bm{r}}[B_{1}^{+};\varepsilon(\bm{r})](\bm{r}) = 0,
\label{general PDE}
\end{equation}
where $\bm{r} \in \mathbb{R}^{d}$ is a spatial coordinate and $\mathcal{N}_{\bm{r}}[\cdot;\varepsilon]$ is a symbolic representation of the Helmholtz equation \eqref{Heq} or \eqref{generalized Heq without Bz}. $\varepsilon(\bm{r})$ denotes the complex-valued EP at the location $\bm{r}$ and $B_{1}^{+}(\bm{r})$ describes the hidden $B_{1}^{+}$ field solution governed by equation \eqref{general PDE}. Given $N$ noisy and/or incomplete measurements $\{\bm{r}_{i}, \tilde{B}_{1}^{+}(\bm{r}_{i})\}_{i=1}^{N}$, we aim to learn the EP distributions $\varepsilon$ as well as the $B_{1}^{+}$ for all $\bm{r}$. To do so, we define a Fourier neural network $\mathcal{B}_{1}^{+}(\bm{r};\bm{\theta_{1}})$, constructed by Gaussian random Fourier features \cite{tancik2020fourier} followed by a fully-connected neural network with a set of weights and biases $\bm{\theta_{1}}$, to represent the complex $B_{1}^{+}$ field. The Gaussian random Fourier features mapping is defined as:
\begin{equation}
\gamma(\boldsymbol{r})=\left[\begin{array}{c}
\cos (\boldsymbol{B} \boldsymbol{r}) \\
\sin (\boldsymbol{B} \boldsymbol{r})
\end{array}\right],
\label{GRFF}
\end{equation}
where each entry in $\boldsymbol{B} \in R^{m \times 3}$ is sampled from a Gaussian distribution $\mathcal{N}(0,s^{2})$. $2m$ equals the width of the fully-connected neural network following the defined Fourier features and $s >0$ is a task-specific hyperparameter. We use an additional fully-connected neural network $\mathcal{E}_{c}(\bm{r};\bm{\theta_{2}})$ with independent weights and biases $\bm{\theta_{2}}$ to estimate the distribution of EP. Hereinafter, we refer to $\mathcal{B}_{1}^{+}(\bm{r};\bm{\theta_{1}})$ and $\mathcal{E}_{c}(\bm{r};\bm{\theta_{2}})$ as $B_{1}^{+}$ net and EP net, respectively. The PDE residual of \eqref{general PDE} is transformed to:
\begin{equation}
\mathcal{R}(\bm{r}, \theta_{1}, \theta_{2}) := \mathcal{N}_{\bm{r}}[\mathcal{B}_{1}^{+}(\bm{r};\bm{\theta_{1}});\mathcal{E}_{c}(\bm{r};\bm{\theta_{2}})](\bm{r}).
\label{residual}
\end{equation}
Similar to other machine learning tasks \cite{wang2023contrastive,chen2021towards}, here a good set of candidate parameters $\{\bm{\theta_{1}, \bm{\theta_{2}}}\}$ can be obtained by minimizing the following composite loss function via gradient descent \cite{ruder2016overview,yang4040704vflh,yang2023gradient} with the Adam optimizer \cite{kingma2014adam}:
\begin{equation}
\small
\begin{aligned}
\mathcal{L}(\bm{\theta_{1},\theta_{2}}) &= \mathcal{L}_{\text{data }}(\bm{\theta_{1}})+\lambda \mathcal{L}_{r}(\bm{\theta_{1}, \theta_{2}}), \\
\mathcal{L}_{\text {data }}(\bm{\theta_{1}}) &=  \frac{1}{N}\sum_{i=1}^{N} |\mathrm{Re}\{\mathcal{B}_{1}^{+}(\bm{r}_{i};\bm{\theta_{1}})\} - \mathrm{Re}\{\tilde{B}_{1}^{+}(\bm{r_{i}})\}|^{2}  \\
&+ \frac{1}{N}\sum_{i=1}^{N}|\mathrm{Im}\{\mathcal{B}_{1}^{+}(\bm{r}_{i};\bm{\theta_{1}})\} - \mathrm{Im}\{\tilde{B}_{1}^{+}(\bm{r_{i}})\}|^{2},\\
\mathcal{L}_{r}(\bm{\theta_{1}, \theta_{2}}) &=\frac{1}{N}\sum_{i=1}^{N} |\mathcal{R}(\bm{r}_{i}, \bm{\theta_{1}, \theta_{2}})|^{2}.
\end{aligned}
\label{composite loss}
\end{equation}
$\mathcal{L}_{\text{data }}$ denotes the data mismatch and $\mathcal{L}_{r}$ denotes the PDE residual. $\lambda$ denotes the weight coefficient in the loss function, which balances the two loss terms in the composite loss. We remark that $\lambda$ is a hyperparameter that can either be specified by the user or be tuned automatically \cite{wang2022and, wang2021understanding}. All the derivatives of $\mathcal{B}_{1}^{+}(\bm{r};\bm{\theta_{1}})$ and $\mathcal{E}_{c}(\bm{r};\bm{\theta_{2}})$ with respect to the spatial coordinate $\bm{r}$ as well as the gradient of the loss function with respect to the neural network parameters $\{\bm{\theta_{1}, \bm{\theta_{2}}}\}$, are computed using automatic differentiation algorithms \cite{baydin2018automatic}.

The workflow of our proposed PIFON-EPT (Fig. \ref{fig:PIFON}) can be summarized as follow. First, we define two separate fully-connected neural networks $B_{1}^{+}$ Net and EP Net ($\mathcal{E}_{c}(\bm{r};\bm{\theta_{2}})$) to represent the $B_{1}^{+}$ and the EP, respectively. A random Fourier features mapping (see Fig. \ref{fig:PIFON} green dotted box) is embedded into $B_{1}^{+}$ Net to learn high frequency components of the target $B_{1}^{+}$ field solution more efficiently \cite{tancik2020fourier}. Second, $B_{1}^{+}$ Net and EP Net are trained jointly by minimizing a composite loss function that aims to fit the measured $\tilde{B}_{1}^{+}$ data (see Fig. \ref{fig:PIFON} blue dotted box), while satisfying the physics laws characterized by the PDE residual. The trained physics-informed $B_{1}^{+}$ Net and EP Net facilitate the generation of physically consistent $B_{1}^{+}$ and EP predictions at any desired spatial point, respectively (see Fig. \ref{fig:PIFON} bottom red dotted boxes). In particular the $B_{1}^{+}$ Net denoises and completes the input $\tilde{B}_{1}^{+}$.

\subsection{Choice of Helmholtz equation}
If we assume piece-wise constant EP, then the Helmholtz equation simplifies as in \eqref{Heq}. Eq. \eqref{generalized Heq without Bz} is a generalized form of the same equation, which accounts for gradients of EP, but is yet not fully general because to reduce the number of unknowns, we assumed that $B_{z}$ is equal to zero. Depending on which Helmholtz equation is used, we introduced two variants of PIFON-EPT: \textit{simplified PIFON-EPT} and \textit{generalized PIFON-EPT}. 

\subsubsection{Simplified PIFON-EPT} Assumes piece-wise constant EP and does not require any assumption on $B_{z}$. Following Eq. \eqref{Heq}, the Helmholtz residual \eqref{residual} can be represented as:
\begin{equation}
\mathcal{R}_{H} = \nabla^{2}\mathcal{B}_{1}^{+}(\bm{r};\bm{\theta_{1}}) + k_{0}^{2}\mathcal{E}_{c}(\bm{r};\bm{\theta_{2}})\mathcal{B}_{1}^{+}(\bm{r};\bm{\theta_{1}}).
\label{H residual}
\end{equation}

\subsubsection{Generalized PIFON-EPT} Assumes $B_{z} \approx 0$ and uses the generalized equation \eqref{generalized Heq without Bz}. The Helmholtz residual \eqref{residual} becomes:
\begin{equation}
\footnotesize
\begin{aligned}
&\mathcal{R}_{GH} = \nabla^{2}\mathcal{B}_{1}^{+}(\bm{r};\bm{\theta_{1}}) + k_{0}^{2}\mathcal{E}_{c}(\bm{r};\bm{\theta_{2}})\mathcal{B}_{1}^{+}(\bm{r};\bm{\theta_{1}}) - \frac{1}{\mathcal{E}_{c}(\bm{r};\bm{\theta_{2}})} \\
&  \left(\frac{\partial \mathcal{B}_1^{+}(\bm{r};\bm{\theta_{1}})}{\partial x}-i \frac{\partial \mathcal{B}_1^{+}(\bm{r};\bm{\theta_{1}})}{\partial y} \right)\left(\frac{\partial  \mathcal{E}_{c}(\bm{r};\bm{\theta_{2}})}{\partial x} +i \frac{\partial \mathcal{E}_{c}(\bm{r};\bm{\theta_{2}})}{\partial y} \right)  \\
&- \frac{1}{\mathcal{E}_{c}(\bm{r};\bm{\theta_{2}})} \left(\frac{\partial \mathcal{B}_1^{+}(\bm{r};\bm{\theta_{1}})}{\partial z}\cdot\frac{\partial  \mathcal{E}_{c}(\bm{r};\bm{\theta_{2}})}{\partial z}\right) .
\end{aligned}
\label{GH residual}
\end{equation}

Both techniques rely on knowledge of the absolute phase of $B_{1}^{+}$, which for a quadrature birdcage coil can be estimated from the transceive phase assumption. Note that with a sufficient number of transmit-receive coils, it is theoretically possible to solve for both the unknown absolute phase and $B_{z}$ \cite{SodicksonGeneralizedLMT2013}, although the lack of suitable multi-channel coils and the computational complexity of such solution has prevented practical implementations.

\section{Results} \label{Results}
We present a series of numerical examples to demonstrate the effectiveness of our proposed PIFON-EPT framework. Throughout all experiments, unless otherwise specified, we used simulated complex $B_{1}^{+}$ maps as measured data and corrupted them with white Gaussian noise with a standard deviation equal to the ratio of the peak value of $|B_{1}^{+}|$ to a prescribed peak signal-to-noise-ratio (SNR) value. The simulations were performed with the volume \cite{giannakopoulos2019memory} and the volume\hyp surface integral equation \cite{giannakopoulos2021tensor, giannakopoulos2022hybrid} methods. The volume equations were solved using higher-order polynomials \cite{georgakis2020fast} as basis functions to ensure accuracy in the $B_{1}^{+}$ distributions. All experiments were performed on a server running Ubuntu 20.04.3 LTS operating system, with an Intel(R) Xeon(R) Silver 4216 CPU at 2.10GHz, 64 cores, 2 threads per core, and an NVIDIA RTX 3090 GPU with 24 GB of memory. 

To measure the discrepancy between the prediction ($\hat{\bm{v}}$) and ground-truth ($\bm{v} \in \mathbb{R}^{N}$) values we used the peak normalized absolute error (PNAE), defined as:
\begin{equation}
\text{PNAE}(\hat{\bm{v}},\bm{v}) = \frac{\|\bm{v}-\hat{\bm{v}}\|_{1}}{\|\bm{v}\|_{\infty}}.
\end{equation}

\subsection{Validation against the analytical solution} \label{Mie theory example}
To verify our method, we used a complex $B_{1}^{+}$ map obtained from the Mie Scattering theory \cite{wiscombe1980improved} for an infinitely long homogeneous dielectric cylinder of relative permittivity $3$ and electric conductivity $0.01$ S/m, and it's air outside the cylinder. The operating wavelength was $\lambda = 2.437$ m and the cylinder had a radius $r$ equal to the wavelength. A TMz planewave was used as the excitation.

\subsubsection{Data Acquisition}
We considered a representative section of the cylinder and computed the $B_{1}^{+}$ field distribution in the domain $[-2r, 2r] \times [-2r, 2r]$ using Mie scattering theory \cite{bohren2008absorption}. The pixel isotropic resolution was set to $0.05 \lambda$ so that the section was $81 \times 81$ for a total of 6561 voxels. We corrupted the synthetic $B_{1}^{+}$ field with Gaussian noise of peak SNR of 200 and then scaled the noisy field with the peak value of $|B_{1}^{+}|$ to obtain synthetic $\tilde{B}_{1}^{+}$ measurements. The resulting $\tilde{B}_{1}^{+}$ fields were used as the measured data for PIFON-EPT.



\subsubsection{PIFON Training Settings}
$B_{1}^{+}$ Net was constructed by a Fourier features mapping initialized with $s = 2$ as a coordinate embedding of the input, followed by a fully-connected neural network with 3 layers, 128 units per layer. EP Net was constructed using a fully-connected neural network with 3 layers, 128 units per layer. We set all the activation functions as the Sine function. We set $\lambda = 10^{-4}$ in equation \eqref{composite loss}. We trained $B_{1}^{+}$ Net and EP Net jointly using the Adam optimizer for 120k iterations in total, with a decaying schedule of learning rates $10^{-3}$, $10^{-4}$, $10^{-5}$ decreased every 40k iterations, which took $\sim 30$ minutes and $\sim 40$ minutes for employing simplified PIFON-EPT and generalized PIFON-EPT, respectively.

\subsubsection{Results}
\begin{figure}[ht!]
    \begin{center}
    \includegraphics[scale=0.18]{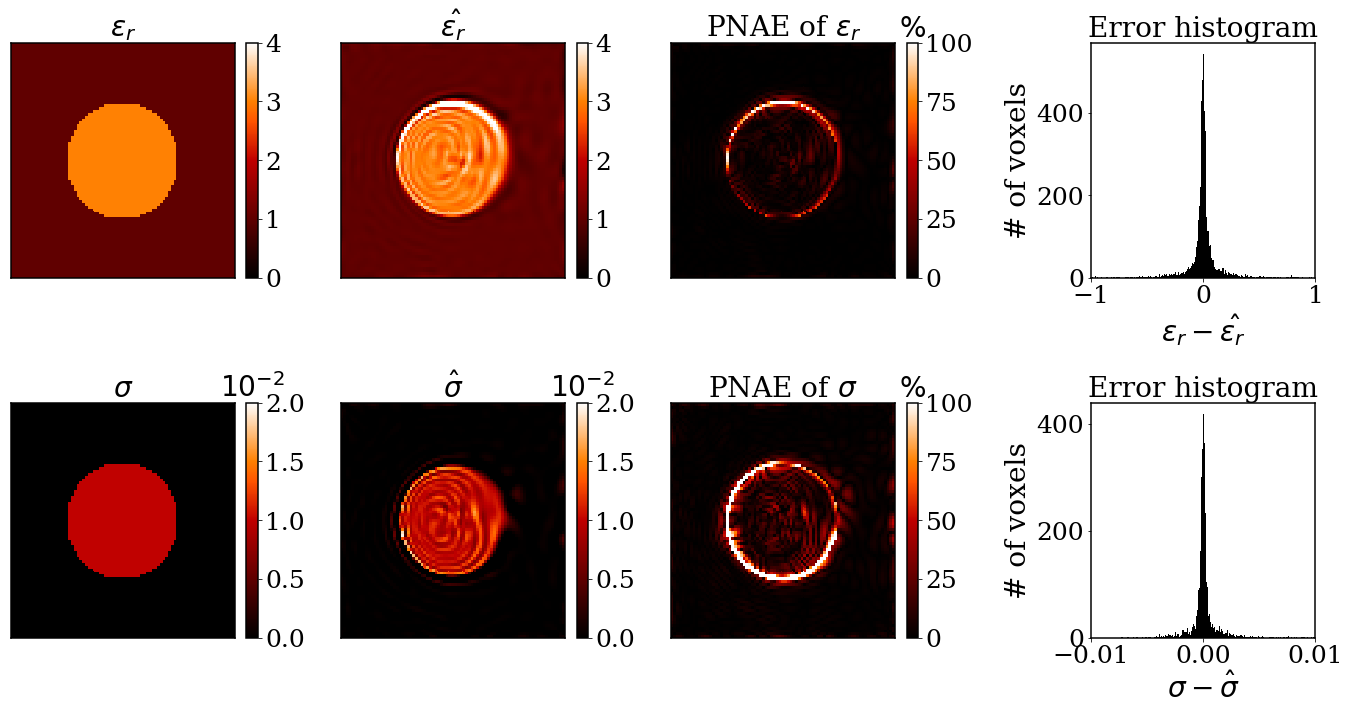}  
    \vspace{-5pt}
    \caption{EP reconstruction with simplified PIFON-EPT for a representative section of the uniform dielectric cylinder. From left to right, ground truth EP, including relative permittivity (top) and conductivity (bottom), predicted EP using $\tilde{B}_{1}^{+}$ measurements with peak SNR of 200, peak-normalized absolute errors, distribution of the error in 6561 voxels.}
    \label{fig:2d nongen EP results}
    \end{center}
\vspace{-10pt}
\end{figure}

\begin{figure}[ht!]
    \begin{center}
    \includegraphics[scale=0.18]{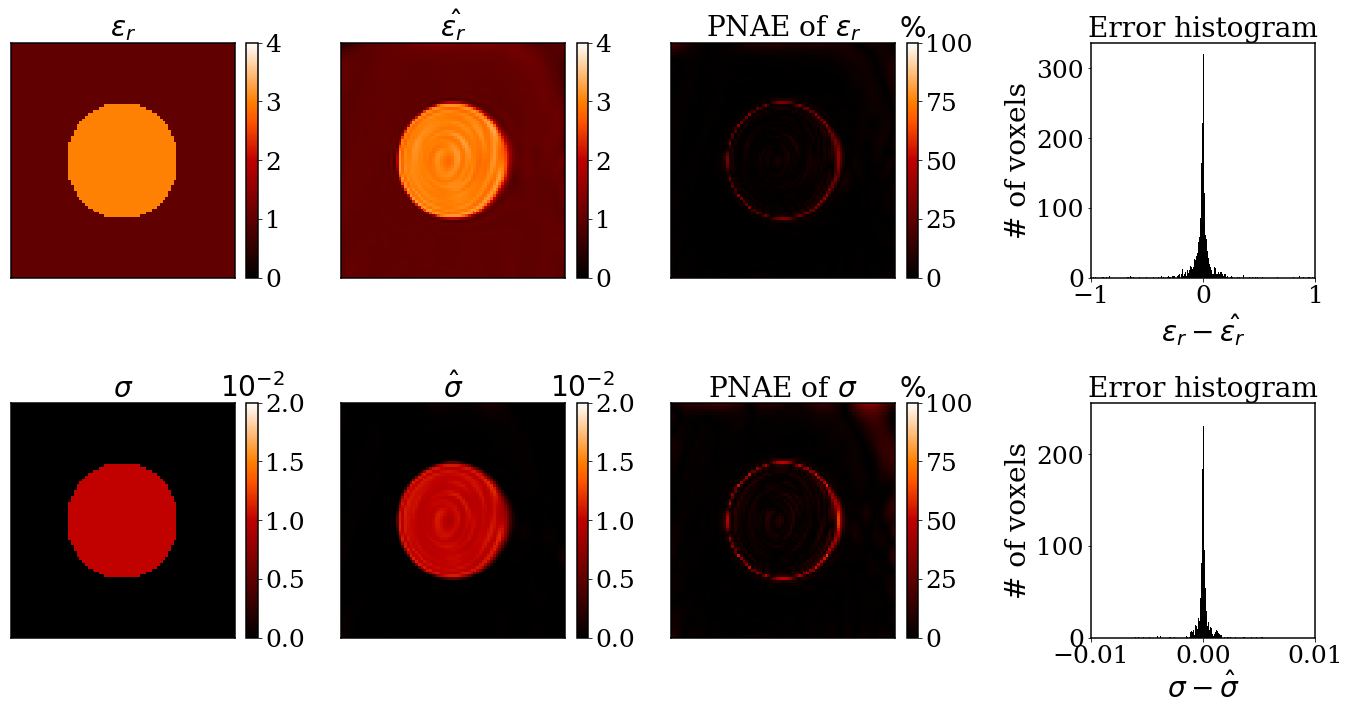}  
    \vspace{-5pt}
    \caption{EP reconstruction with generalized PIFON-EPT for a representative section of the uniform dielectric cylinder. From left to right, ground truth EP, including relative permittivity (top) and conductivity (bottom), predicted EP using $\tilde{B}_{1}^{+}$ measurements with peak SNR of 200, peak-normalized absolute errors, distribution of the error in 6561 voxels.}
    \label{fig:2d gen EP results}
    \end{center}
\vspace{-5pt}
\end{figure}

\begin{figure}[ht!]
    \begin{center}
    \includegraphics[scale=0.18]{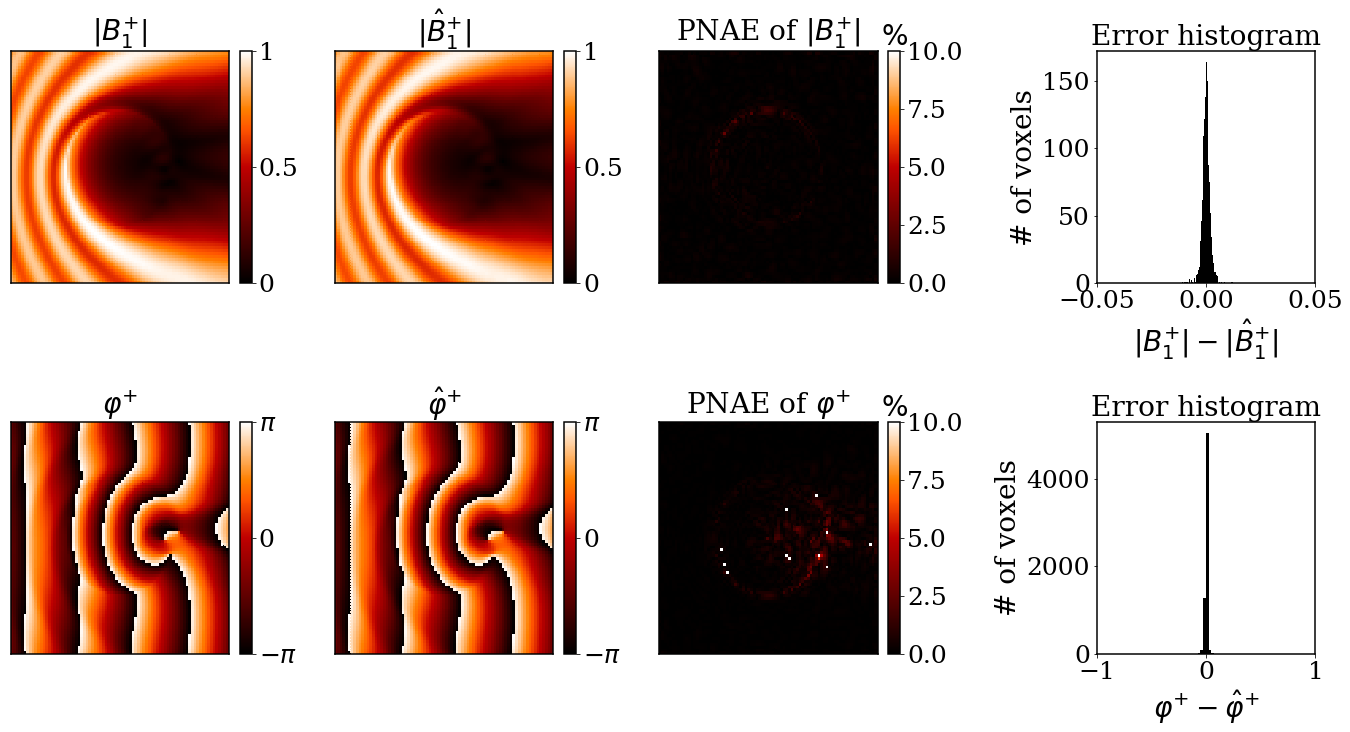}  
    \vspace{-5pt}
    \caption{Reconstructed $B_{1}^{+}$ with simplified PIFON-EPT for a representative section of the uniform dielectric cylinder. From left to right, ground truth noise-free synthetic $B_{1}^{+}$, including magnitude (top) and transmit phase (bottom), reconstructed $B_{1}^{+}$ from noisy synthetic $\tilde{B}_{1}^{+}$ measurements with peak SNR of 200, peak-normalized absolute errors, distribution of the error in 6561 voxels.}
    \label{fig:2d nongen b1 results}
    \end{center}
\vspace{-5pt}
\end{figure}

\begin{figure}[ht!]
    \begin{center}
    \includegraphics[scale=0.18]{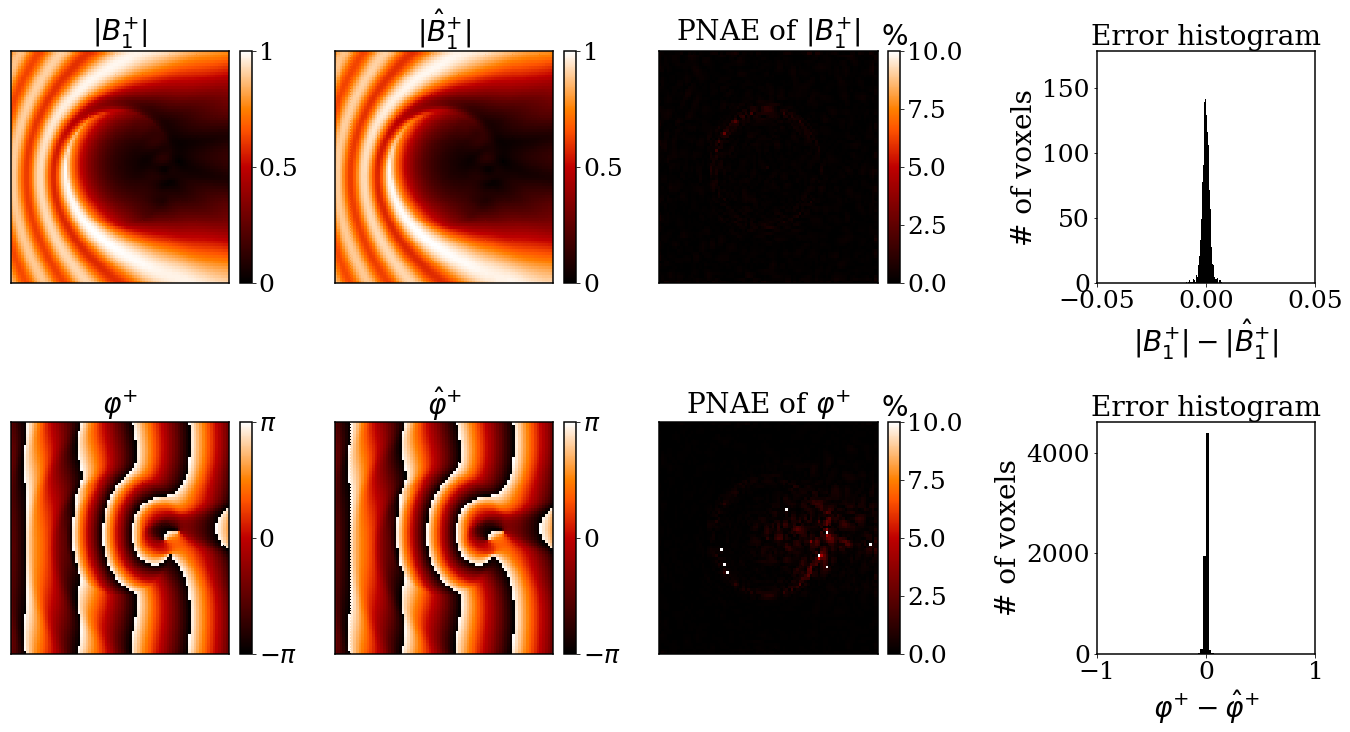}  
    \vspace{-5pt}
    \caption{Reconstructed $B_{1}^{+}$ with generalized PIFON-EPT for a representative section of the uniform dielectric cylinder. From left to right, ground truth noise-free synthetic $B_{1}^{+}$, including magnitude (top) and transmit phase (bottom), reconstructed $B_{1}^{+}$ from noisy synthetic $\tilde{B}_{1}^{+}$ measurements with peak SNR of 200, peak-normalized absolute errors, distribution of the error in 6561 voxels.}
    \label{fig:2d gen b1 results}
    \end{center}
\vspace{-10pt}
\end{figure}

We tested the performance of the simplified and generalized PIFON-EPT using the same training settings. Fig. \ref{fig:2d nongen EP results} and Fig. \ref{fig:2d gen EP results} compare the reconstructed EP against the ground truth values for the simplified and generalized PIFON-EPT, respectively. Fig. \ref{fig:2d nongen b1 results} and Fig. \ref{fig:2d gen b1 results} compare ground truth and reconstructed $B_{1}^{+}$ maps for the simplified and generalized PIFON-EPT, respectively. The average PNAE over the domain for the relative permittivity, conductivity, and $B_{1}^{+}$ was $3.96\%$, $9.67\%$ and $0.22\%$, respectively for the simplified PIFON-EPT. The error decreased to $1.80\%$, $1.11\%$ and $0.20\%$, when the generalized PIFON-EPT was used. The lower error in this case is because the generalized PIFON-EPT is able to approximate better EPs at the boundary.

\subsection{Concentric Cylindrical Phantom} \label{birdcage example}
\begin{figure}[t]
\begin{center}
\includegraphics[width=0.3\textwidth, trim={0 0 0 0}]{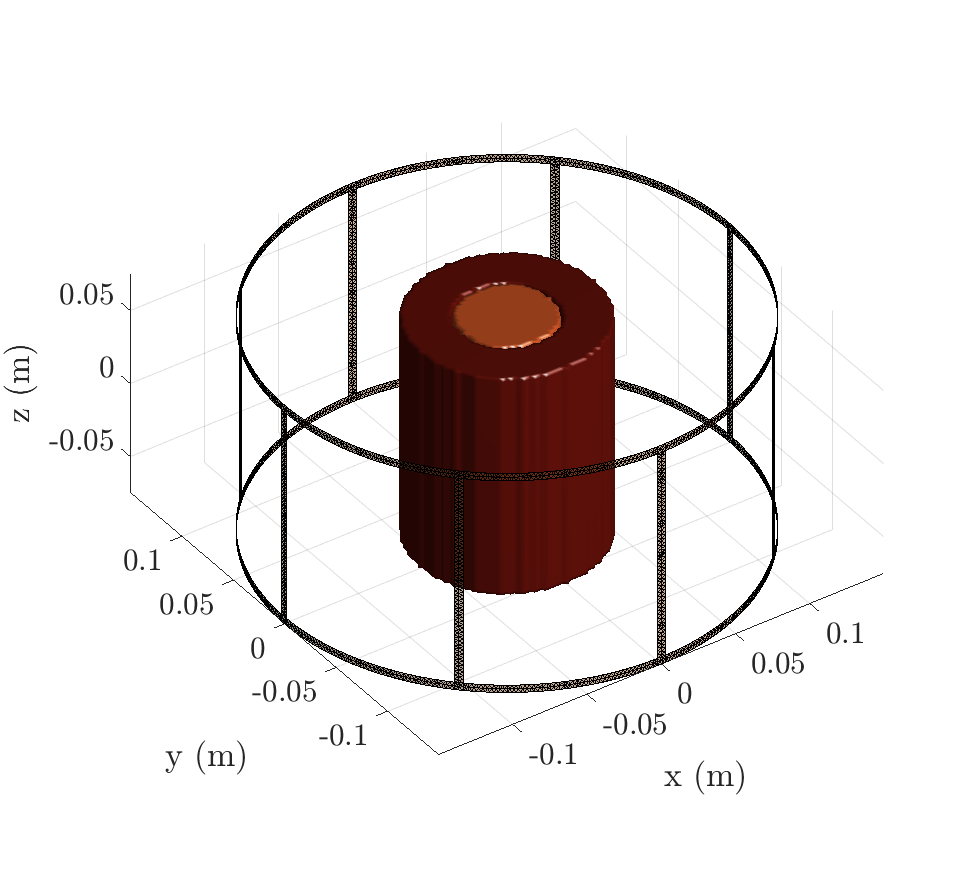}
\vspace{-5pt}
\caption{Geometry of the high-pass birdcage coil loaded with a two-compartments cylindrical phantom}
\label{birdcage}
\end{center}
\vspace{-10pt}
\end{figure}

We considered a two-compartment concentric cylindrical phantom with relative permittivity $\varepsilon=\{70, 78\}$ and conductivity $\sigma=\{0.5, 1\}$ S/m (outer, inner). The cylinder loaded a high\hyp pass birdcage coil with eight legs as shown in Fig. \ref{birdcage}. The outer and inner radius of the cylinder were 6 cm and 3 cm, respectively, and its length was 14 cm. For this example, we compared the proposed PIFON-EPT with the Helmholtz-EPT (H-EPT) and the Convection-Reaction EPT (CR-EPT) (see \ref{subsec:EPT}). In particular, we used the implementations in EPTlib \cite{arduino2021eptlib}, with the Savitzky-Golay filter with an ellipsoid-shaped kernel of size $2\times 2 \times 2$ to approximate all the gradients. For CR-EPT, we set the diffusion coefficient to $0.02$ and the conductivity boundary condition to $0.55$ S/m. 

\subsubsection{Data Acquisition} We used the volume-surface integral equation method \cite{giannakopoulos2022hybrid} to simulate the circularly polarized (CP) mode of the birdcage coil loaded with the cylindrical phantom at 3 T. The resolution was set to 2 mm$^3$. We used $B_{1}^{+}$ and $B_{1}^{-}$ from the central region of the cylinder (12 $\times$ 12 $\times$ 2 cm$^3$, MR measurements out of cylindrical phantom were not used) and corrupted them with Gaussian noise of peak SNR of 200. We approximated the complex $B_{1}^{+}$ using the transceive phase assumption (TPA) and constructed the MR measurements $|\tilde{B}_{1}^{+}|$ and $\tilde{\varphi}^{\pm}$. 


\subsubsection{PIFON Training Settings} The $B_z$ field of a birdcage is negligible around the mid-plane of the coil. For this reason, we used the generalized PIFON-EPT to perform the reconstruction. For $B_{1}^{+}$ Net, the Fourier feature mapping was initialized with $s = 40$ as a coordinate embedding of the input, followed by a fully-connected neural network with 6 layers, 128 units per layer. EP Net was an additional Fourier neural network constructed by a Fourier feature mapping initialized with $s = 2$, followed by a fully-connected neural network with 6 layers, 128 units per layer. We set all the activation functions as the Sine function and set $\lambda = 10^{-8}$ in equation \eqref{composite loss}. We trained $B_{1}^{+}$ Net and EP Net jointly using the Adam optimizer for 120k iterations in total, with a decaying schedule of learning rates $10^{-3}$, $10^{-4}$, $10^{-5}$ decreased every 40k iterations. Note that the network settings have to change for different experimental setups". In particular, the total number of iterations is determined based on the network size, and deep neural networks usually require more iterations to converge than shallow networks. The overall training time was 220 minutes on our GPU. 

\subsubsection{Results}
\begin{figure}[ht!]
    \begin{center}
    \includegraphics[scale=0.18]{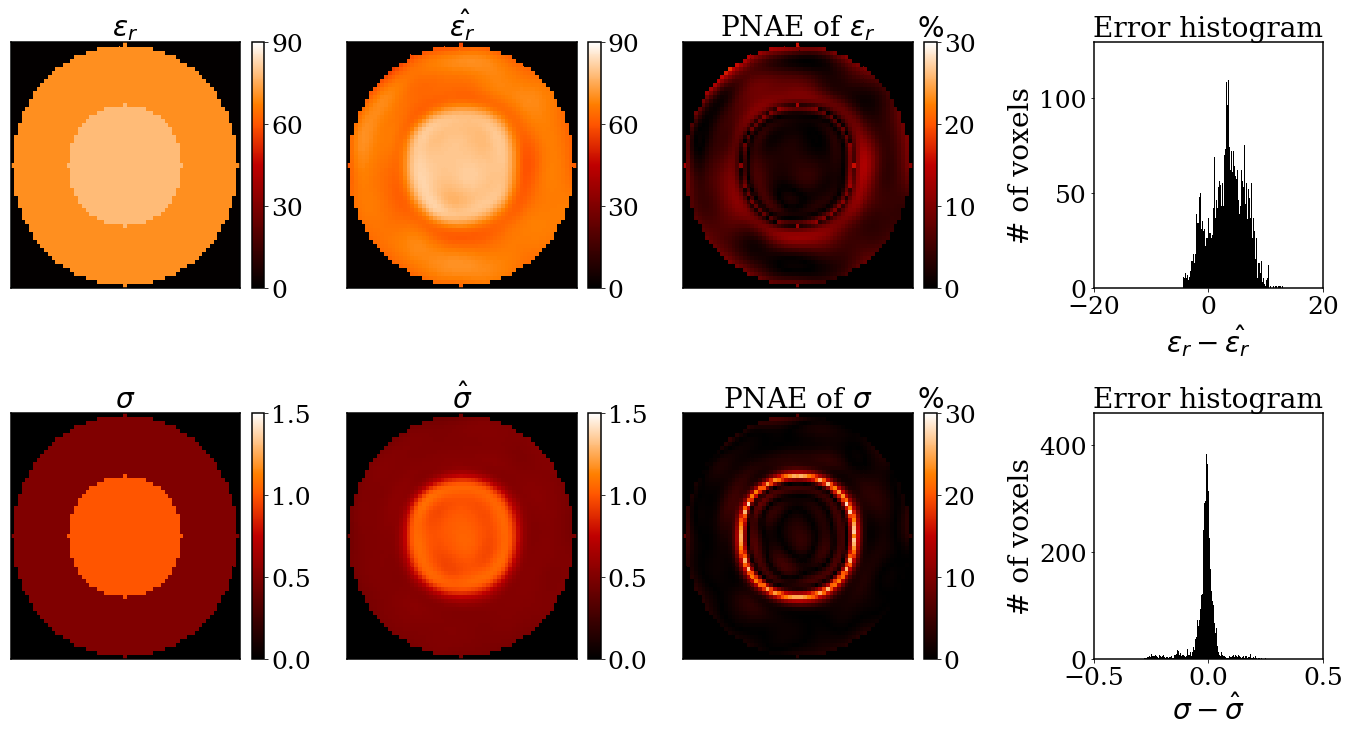} 
    \vspace{-5pt}
    \caption{EP reconstructed with generalized PIFON-EPT for the two-compartment cylindrical phantom. From left to right, ground truth EP for the central axial cut of the phantom, including relative permittivity (top) and conductivity (bottom), estimated EP using synthetic $\tilde{B}_{1}^{+}$ measurements with peak SNR of 200, peak-normalized absolute errors, distribution of the error in 31031 voxels.}
    \label{fig:bird EP results}
    \end{center}
\vspace{-5pt}
\end{figure}

\begin{figure}[ht!]
    \begin{center}
    \includegraphics[scale=0.18]{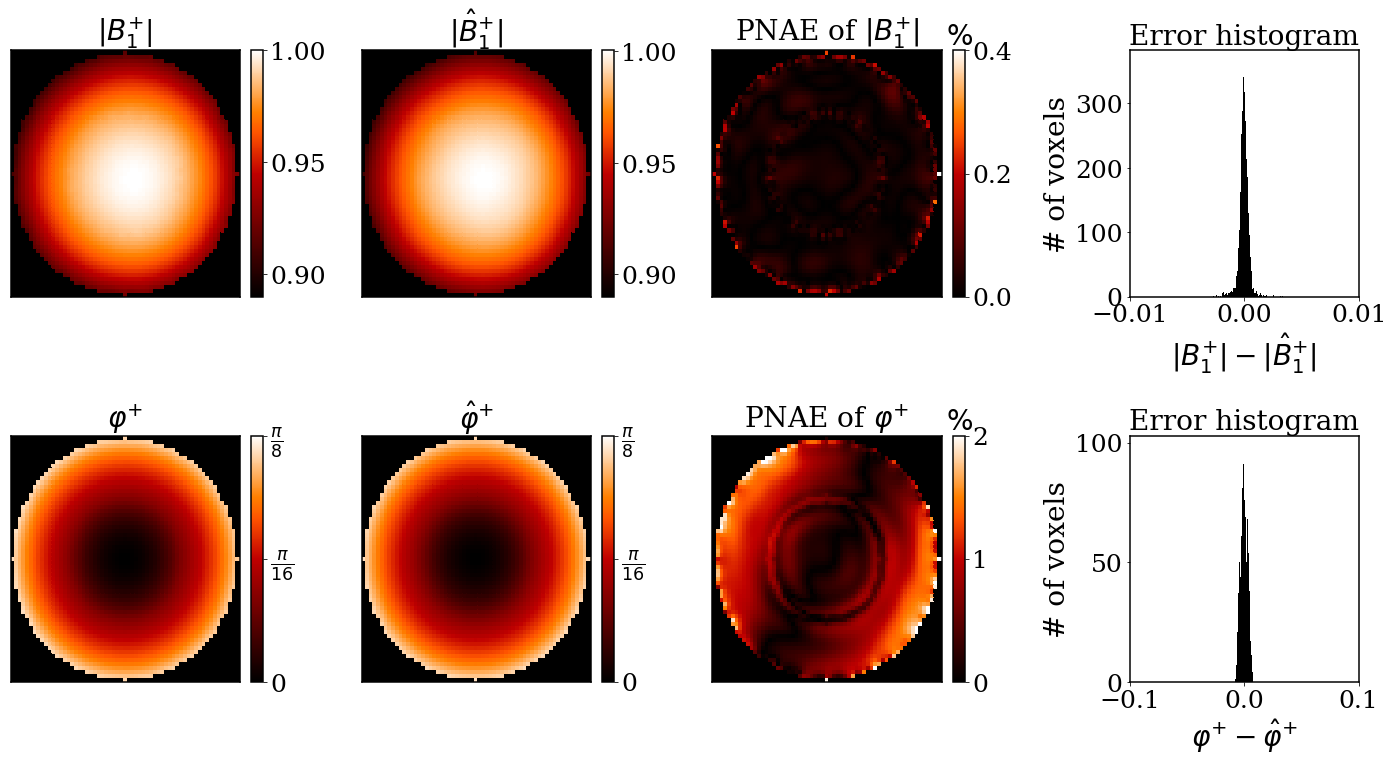}  
    \vspace{-5pt}
    \caption{Reconstructed $B_{1}^{+}$ with generalized PIFON-EPT for the two-compartment cylindrical phantom. From left to right, noise-free synthetic $B_{1}^{+}$ for the central axial cut, including magnitude (top) and transmit phase (bottom), reconstructed $B_{1}^{+}$ field from noisy $\tilde{B}_{1}^{+}$ measurements, peak-normalized absolute errors, distribution of the error in 31031 voxels.}
    \label{fig:bird b1 results}
    \end{center}
\vspace{-5pt}
\end{figure}

 The reconstructed EP (Fig. \ref{fig:bird EP results}) and $B_{1}^{+}$ (Fig. \ref{fig:bird b1 results}) are presented for the central axial cut of the cylinder. The average PNAE over the entire volume of the cylinder was $4.84\%$, $3.20\%$ and $0.25\%$ for the relative permittivity, conductivity and $B_{1}^{+}$, respectively.

\begin{figure}[ht!]
    \begin{center}
    \includegraphics[scale=0.18]{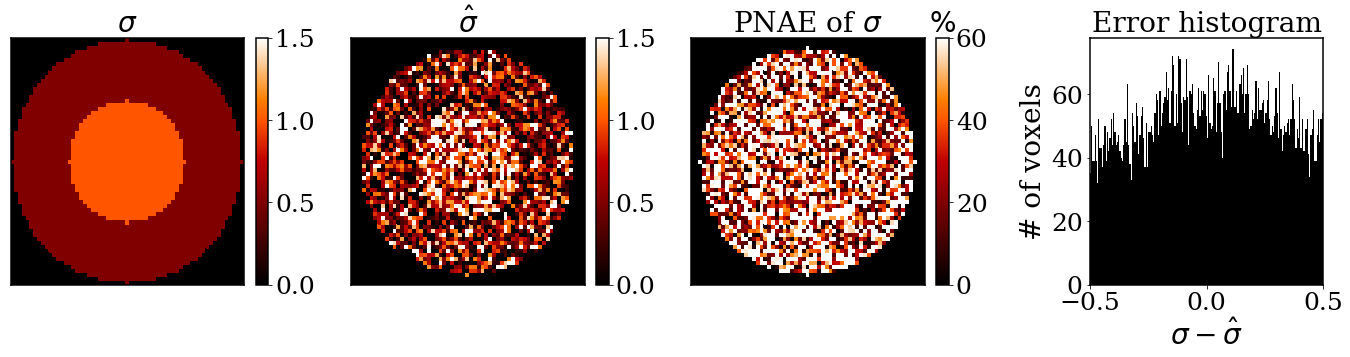} 
    \vspace{-5pt}
    \caption{Conductivity reconstructed with phase-based H-EPT for the two-compartment cylindrical phantom. From left to right, ground truth conductivity for the central axial cut of the phantom, estimated conductivity using $\tilde{\varphi}^{\pm}$ measurements with peak SNR of 200, the peak-normalized absolute errors, the distribution of the error in 17423 voxels.}
    \label{fig:bird EP results H-EPT}
    \end{center}
\vspace{-5pt}
\end{figure}

\begin{figure}[ht!]
    \begin{center}
    \includegraphics[scale=0.18]{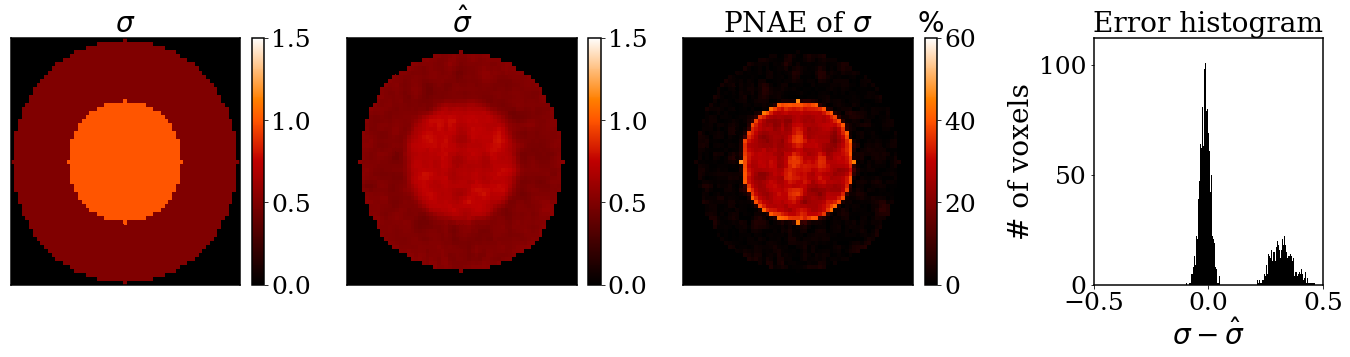} 
    \vspace{-5pt}
    \caption{Conductivity reconstructed with phase-based CR-EPT for the two-compartment cylindrical phantom. From left to right, ground truth conductivity for the central axial cut of the phantom, estimated conductivity using $\tilde{\varphi}^{\pm}$ measurements with peak SNR of 200, the peak-normalized absolute errors, the distribution of the error in 11645 voxels.}
    \label{fig:bird EP results CR-EPT}
    \end{center}
\vspace{-10pt}
\end{figure}

Fig. \ref{fig:bird EP results H-EPT} and Fig. \ref{fig:bird EP results CR-EPT} present the conductivity reconstruction results for H-EPT and CR-EPT, respectively, along with the PNAE distribution and the error histogram. The average PNAE over the volume of the phantom was $51.80\%$ and $11.28\%$ for H-EPT and CR-EPT, respectively.

\subsection{Four-Compartment Phantom} 
In this example, we explore the performance of PIFON-EPT at 7 T. We considered a previously used \cite{serralles2019noninvasive} tissue-mimicking four-compartment phantom shaped as a $20 \times 20 \times 20$ cm$^3$ rectangular parallelepiped. The relative permittivity values of the four compartments were $51$, $56$, $65$, and $76$. The corresponding electric conductivity values were $0.56$, $0.69$, $0.84$, and $1.02$ S/m.

\subsubsection{Data Acquisition} 
\begin{figure}[ht!]
\begin{center}
\includegraphics[width=0.48\textwidth, trim={0 0 0 0}]{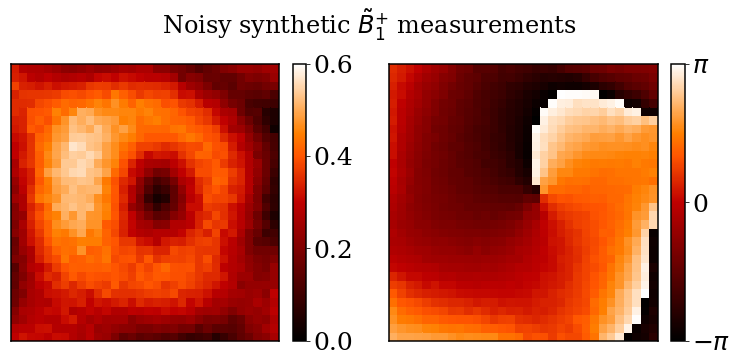}
\vspace{-5pt}
\caption{Noisy synthetic $\tilde{B}_{1}^{+}$ measurements. Magnitude (left) and transmit phase (right) are shown for the central axial cut of the Four-compartment phantom. The peak SNR was set to 50.}
\label{4c data}
\end{center}
\vspace{-10pt}
\end{figure}

We used a single external excitation to illuminate the phantom, generated from a numerical electromagnetic basis \cite{georgakis2022novel}, similar to previous work \cite{serralles2019noninvasive}. We used 6 mm isotropic voxel resolution. We corrupted the synthetic $B_{1}^{+}$ with different levels of Gaussian noise (Peak SNR = 200, 100, 50, 20) and then scaled each field map by the corresponding peak value of $|B_{1}^{+}|$ to obtain synthetic $\tilde{B}_{1}^{+}$ measurements. The case of peak SNR = 50 is shown in Fig. \ref{4c data}. 

\subsubsection{PIFON Training Settings} Since the $B_1$ field in the $z$ direction cannot be assumed zero at $7$ T, we used the simplified PIFON-EPT. The $B_{1}^{+}$ Net was constructed using a Fourier feature mapping initialized with $s = 40$ as a coordinate embedding of the input, followed by a fully-connected neural network with 3 layers, 128 units per layer. For EP Net, we used a second fully-connected neural network with 3 layers, 128 units per layer. We set all the activation functions as the Sine function. We set $\lambda = 10^{-8}$ in equation \eqref{composite loss}. We trained $B_{1}^{+}$ Net and EP Net jointly using the Adam optimizer for 30k iterations in total, with a decaying schedule of learning rates $10^{-3}$, $10^{-4}$, $10^{-5}$ decreased every 10k iterations, which took 21.4 minutes on our GPU.

\subsubsection{Results}
\begin{figure}[t]
    \begin{center}
    \includegraphics[scale=0.18]{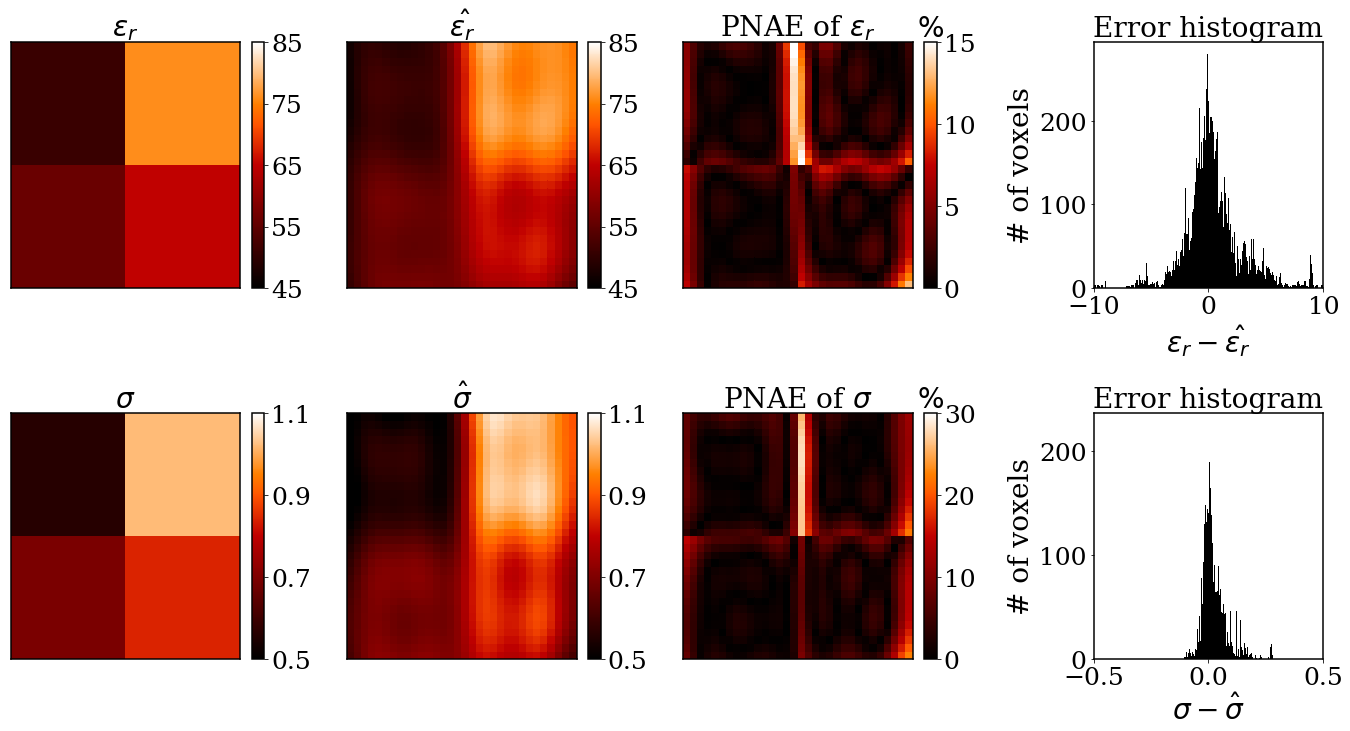}  
    \vspace{-5pt}
    \caption{EP reconstructed with simplified PIFON-EPT for the four-compartment phantom. From left to right, ground truth EP for the central axial cut of the phantom, including relative permittivity (top) and conductivity (bottom), EP reconstructed from synthetic $\tilde{B}_{1}^{+}$ measurements with peak SNR of 50, peak-normalized absolute errors, error distribution in 32768 voxels.}
    \label{fig:4c EP results}
    \end{center}
\vspace{-5pt}
\end{figure}

\begin{figure}[t]
    \begin{center}
    \includegraphics[scale=0.18]{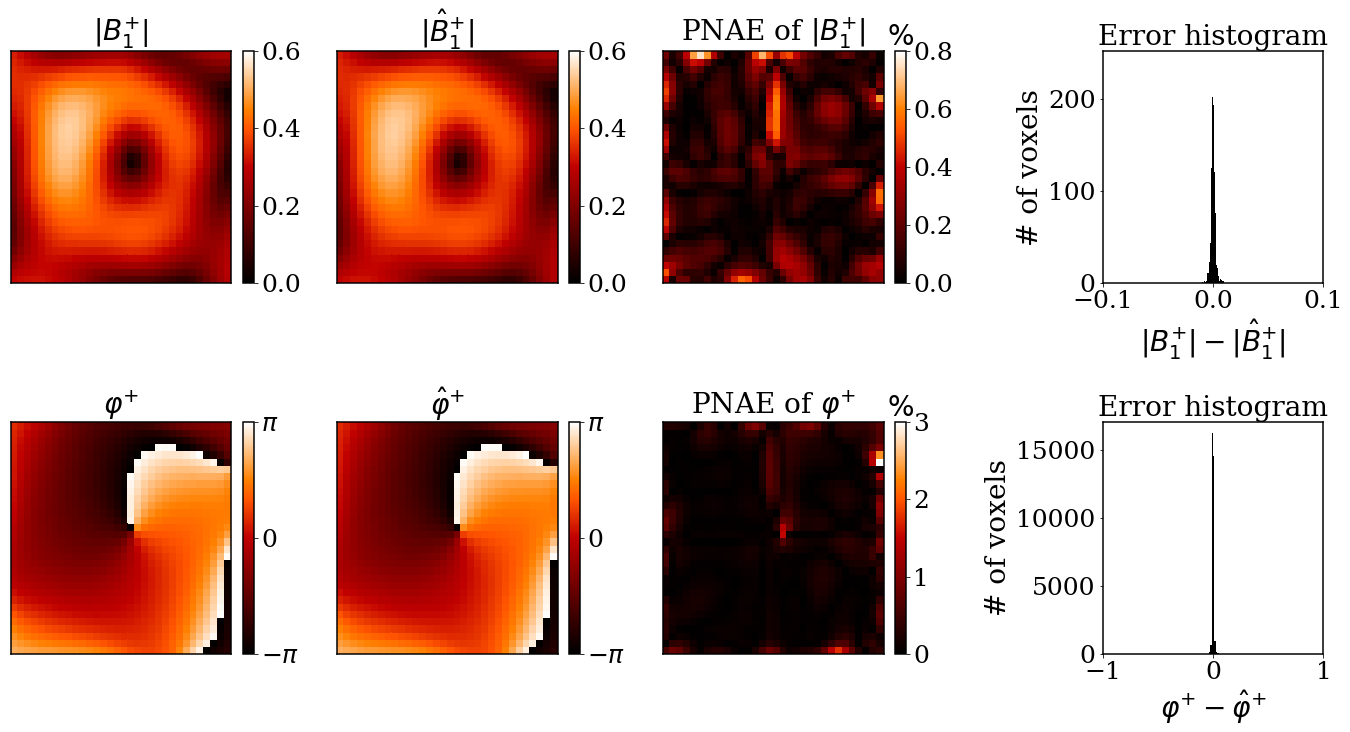}  
    \vspace{-5pt}
    \caption{Reconstructed $B_{1}^{+}$ with simplified PIFON-EPT for the four-compartment phantom. From left to right, ground truth synthetic $B_{1}^{+}$ for the central axial cut of the phantom, including magnitude (top) and transmit phase (bottom), reconstructed $B_{1}^{+}$ field from noise-corrupted synthetic $\tilde{B}_{1}^{+}$ measurements with peak SNR of 50, the peak-normalized absolute errors, the distribution of the error in 32768 voxels.}
    \label{fig:4c b1 results}
    \end{center}
\vspace{-10pt}
\end{figure}

Figs. \ref{fig:4c EP results} and \ref{fig:4c b1 results} presents the reconstructed EP and $B_{1}^{+}$ map (absolute value and phase) for the central slice of the four-compartment phantom, respectively. Our method removed the noise from the noisy synthetic measurements (Fig. \ref{4c data}) and the reconstructed $B_{1}^{+}$ (Fig. \ref{fig:4c b1 results}) was indistinguishable from the noise-free ground truth. The average PNAE over the volume of the phantom was $2.47\%$, $4.01\%$, $0.24\%$ for the relative permittivity, conductivity and $B_{1}^{+}$, respectively.

The average PNAE for the reconstructed EP and $B_{1}^{+}$ for different levels of noise in the synthetic measurements are summarized in TABLE \ref{tab: Robustness analysis}. The reconstructions were robust for a wide range of noise levels.

\begin{table}[ht]
\centering
\caption{Robustness analysis of PIFON-EPT with respect to the noise level}
{\def\arraystretch{1.3}\tabcolsep=5.5pt
\begin{tabular}{|c|c|c|c|c|}
\hline
\diagbox{PNAE}{Peak SNR}&200&100&50&20\\
\hline 
$\varepsilon_{r}$ & 2.56$\%$ & 2.64$\%$ & 2.47$\%$ & 2.56$\%$  \\
\hline 
$\sigma$ & 4.00$\%$ & 4.10$\%$ & 4.01$\%$ & 3.96$\%$  \\
\hline 
$B_{1}^{+}$ & 0.15$\%$ & 0.17$\%$ & 0.24$\%$ & 0.49$\%$  \\
\hline 
\end{tabular}

\label{tab: Robustness analysis}
}
\end{table}

\subsection{Incomplete Four-Compartment Phantom}
In this final numerical experiment, we used the same four-compartment phantom as before, but we assumed the synthetic $\tilde{B}_{1}^{+}$ measurements were incomplete, which could happen in reality if the measured MR signal used to reconstruct the ${B}_{1}^{+}$ maps is too low or corrupted for certain voxels. We tested whether PIFON-EPT could reconstruct the EP and a complete, denoised $B_{1}^{+}$ for the entire volume.

\subsubsection{Data Acquisition} 
\begin{figure}[t]
\begin{center}
\includegraphics[width=0.48\textwidth, trim={0 0 0 0}]{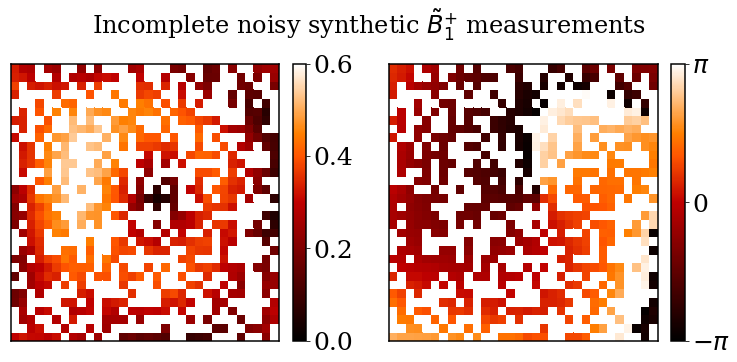}
\vspace{-5pt}
\caption{Incomplete noisy synthetic $\tilde{B}_{1}^{+}$ measurements with $50\%$ of the voxels set to zero. Magnitude (left) and transmit phase (right) are shown for the central axial cut of the Four-compartment phantom. The peak SNR was set to 50.}
\label{in 4c data}
\end{center}
\vspace{-10pt}
\end{figure}

We randomly set to zero from $20\%$ to $90\%$ of the voxels in the synthetic $\tilde{B}_{1}^{+}$ measurements with peak SNR of 50. As a result, only $10\%$ to $80\%$ of the measurements were used as input for simplified PIFON-EPT. Fig. \ref{in 4c data} shows one of the resulting $\tilde{B}_{1}^{+}$ measurements for the central axial cut, where $50\%$ of the $\tilde{B}_{1}^{+}$ values were set to zero.

\subsubsection{Results}
\begin{figure}[t]
    \begin{center}
    \includegraphics[scale=0.18]{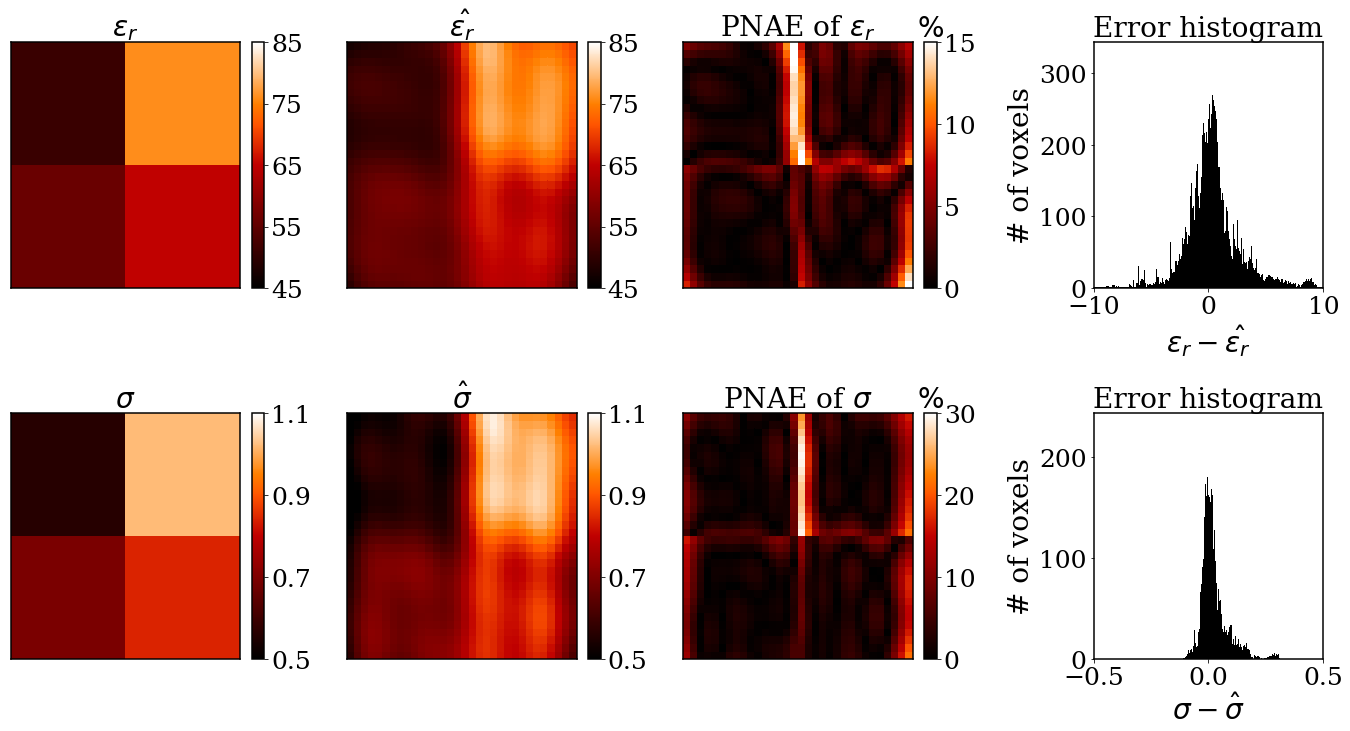} 
    \vspace{-5pt}
    \caption{Reconstructed EP with simplified PIFON-EPT for the incomplete four-compartment phantom. From left to right, ground truth EP for the central axial cut of the phantom, including relative permittivity (top) and conductivity (bottom), estimated EP using $50\%$ of $\tilde{B}_{1}^{+}$ with peak SNR of 50, the peak-normalized absolute errors, the distribution of the error in 32768 voxels.}
    \label{fig:in 4c EP results}
    \end{center}
\vspace{-5pt}
\end{figure}

\begin{figure}[t]
    \begin{center}
    \includegraphics[scale=0.18]{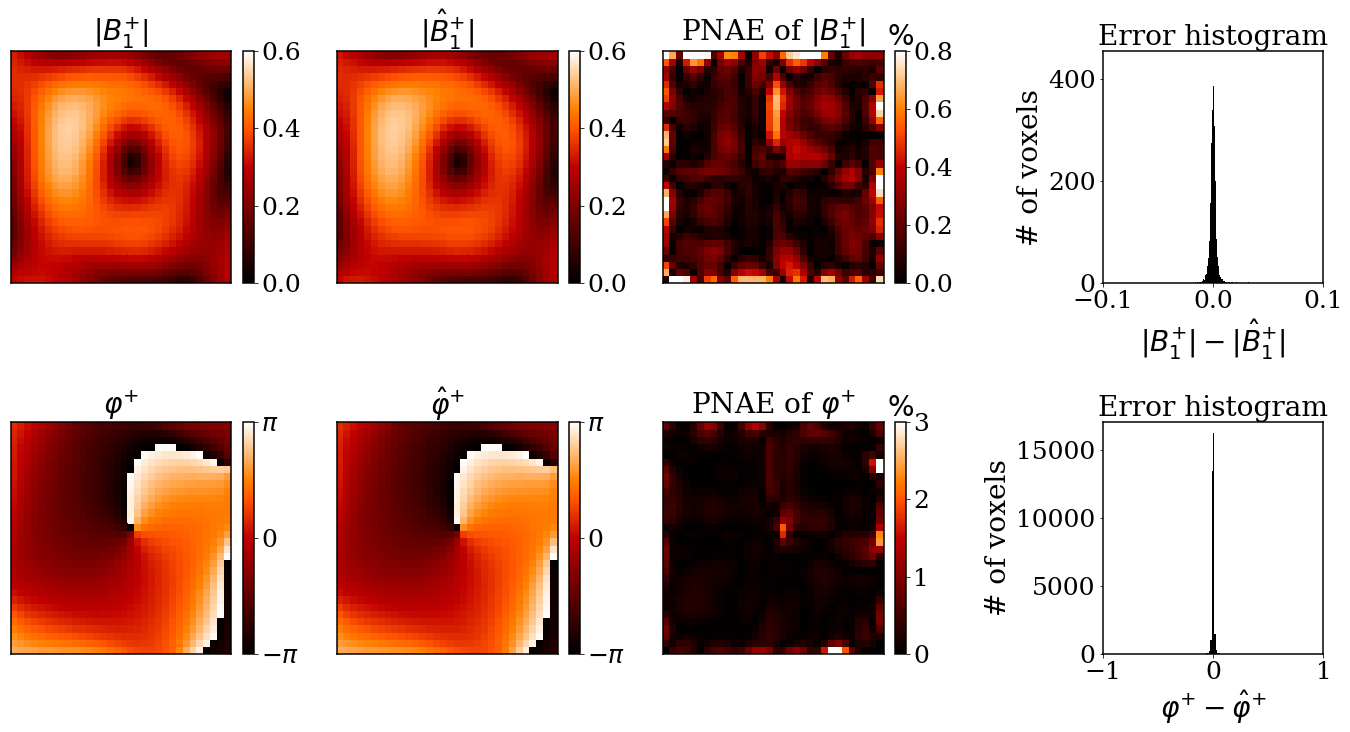}  
    \vspace{-5pt}
    \caption{Reconstructed $B_{1}^{+}$ with simplified PIFON-EPT for the incomplete four-compartment phantom. From left to right, magnitude (top) and transmit phase (bottom) of the synthetic $B_{1}^{+}$ field for the central axial cut of the phantom, reconstructed $B_{1}^{+}$ field using $50\%$ of $\tilde{B}_{1}^{+}$ with peak SNR of 50, peak-normalized absolute errors, error distribution in 32768 voxels.}
    \label{fig:in 4c b1 results}
    \end{center}
\vspace{-10pt}
\end{figure}

We used the same training settings as for the previous experiment. The total training time when we used $10\%$, $20\%$, $50\%$, and $80\%$ of the measurements was $10$, $11$, $15$, and $18$ minutes, respectively. For the case where only $50\%$ of the synthetic $\tilde{B}_{1}^{+}$ measurements were used, Figs. \ref{fig:in 4c EP results} and \ref{fig:in 4c b1 results} show the ground truth EP and noise-free synthetic $B_{1}^{+}$ (first column), the reconstructed EP and the denoised and completed $B_{1}^{+}$ (second column), and the PNAE of the predicted EP and $B_{1}^{+}$ (third column) for the central slice of the phantom. The fourth column presents the error distribution over the entire volume of the phantom. We found that our method could accurately reconstruct the EP and $B_{1}^{+}$ for the whole domain, despite using partial measurements as the input. The average PNAE over the entire volume of the phantom was $2.49\%$, $4.09\%$ and $0.32\%$ for the relative permittivity, conductivity, and $B_{1}^{+}$, respectively.

TABLE \ref{tab: Error analysis} summarizes the average PNAE for the EP and $B_{1}^{+}$ when different percentages of the synthetic measurements were used. The error for the $B_{1}^{+}$ reconstruction increased when a smaller percentage of the data was used. However, PIFON-EPT yielded robust results for the EP maps until as little as $20\%$ of the measurements were used as inputs.

\begin{table}[ht!]
\centering
\caption{Performance of PIFON-EPT with respect to the percentage of measurements used as input for the reconstructions}
{\def\arraystretch{1.3}\tabcolsep=5.5pt
\begin{tabular}{|c|c|c|c|c|}
\hline
\diagbox{PNAE}{$\%$ of the Data}&80$\%$&50$\%$&20$\%$& 10$\%$\\
\hline 
$\varepsilon_{r}$ & 2.41$\%$ & 2.49$\%$ & 2.77$\%$ & 7.22$\%$  \\
\hline 
$\sigma$ & 3.94$\%$ & 4.09$\%$ & 4.06$\%$ & 7.58$\%$  \\
\hline 
$B_{1}^{+}$ & 0.26$\%$ & 0.32$\%$ & 0.57$\%$ & 2.69$\%$  \\
\hline 
\end{tabular}
\label{tab: Error analysis}
}
\end{table}

\section{Discussion} \label{Discussion}
In this work, we reformulated EPT as a physics-constrained optimization problem with the goal to train two independent neural networks ($B_{1}^{+}$ Net and EP Net) to represent the $B_{1}^{+}$ and EP at any location of interest. To achieve that, we minimized a composite loss that aims to fit $\tilde{B}_{1}^{+}$ measurements while penalizing the PDE residual (see Fig. \ref{fig:PIFON}) via gradient descent with Adam optimizer \cite{kingma2014adam}. Penalizing the PDE residual not only helps EP Net predict the EP distributions that best describe the measured data but also prevents $B_{1}^{+}$ Net from fitting the noise. Compared with standard EPT methods \cite{katscher2009determination, hafalir2014convection} that rely on numerical derivatives to approximate gradients of noisy $\tilde{B}_{1}^{+}$ measurements, which is prone to noise amplifications and artifacts, PIFON-EPT uses automatic differentiation \cite{baydin2018automatic} to calculate all the necessary gradients from de-noised $B_{1}^{+}$ maps provided by $B_{1}^{+}$ Net. This way of computing derivatives makes our method robust to noise. Unlike previous supervised deep learning-based EPT methods \cite{mandija2019opening, hampe2020investigating, gavazzi2020deep, giannakopoulosusage, leijsen2022combining}, our approach does not require a large amount of known data pairs to supervise the training. Compared with previous hybrid deep learning EPT methods \cite{inda2022physics1,inda2022physics2}, which combine deep learning and CR-EPT to solve EP from convection-reaction equations, our method directly trains a neural network (EP Net) to represent the EP based on measured data and the Helmholtz PDE without requiring any boundary conditions and hyperparameter tuning for the diffusion coefficient.

A major concern for $B_{1}^{+}$ maps represented by neural networks is that deep fully-connected networks could fail to learn high-frequency components of the target functions because of the spectral bias \cite{ronen2019convergence, cao2019towards, rahaman2019spectral, wang2021eigenvector}. To overcome the spectral bias and ensure that $B_{1}^{+}$ Net would efficiently learn the high-frequency details of $B_{1}^{+}$, we applied Fourier features mapping as an input embedding to the $B_{1}^{+}$. In the concentric cylindrical phantom example, we also applied Fourier features mapping to EP Net because it could help the network avoid predicting homogeneous EP distributions. 

In simplified PIFON-EPT, we assume a homogeneous distribution of EP. This assumption introduces errors near the interface between regions of different EP values and can deteriorate the quality of the reconstructions. When $B_{z}$ is negligible, the generalized PIFON-EPT can be used which allows the estimation of inhomogeneous EP distributions based on the generalized Helmholtz equation \eqref{generalized Heq without Bz} which can greatly decrease the errors near the tissue boundaries (see \ref{Mie theory example}). In fact, we showed that PIFON-EPT returned $48.6 \%$ and $8.08 \%$ more accurate results on average compared to H-EPT and CR-EPT (see \ref{birdcage example}). Furthermore, CR-EPT required tuning of the boundary condition value and the diffusion coefficient parameter until the reconstructed conductivity was close to the ground-truth value, which is not practical in experiments where the ground-truth values are unknown.

To the best of our knowledge, PIFON-EPT is the only EPT method that can reconstruct EP and $B_{1}^{+}$ for an entire object, using incomplete and noisy $B_{1}^{+}$ measurements. We demonstrated this for an ultra-high field MRI example, using complex-valued synthetic $B_{1}^{+}$ measurements. The same approach would be impractical in actual experiments because the absolute phase of the $B_{1}^{+}$ is not measurable and the TPA does not hold at 7 T. However, note that PIFON-EPT could be adapted to work with multiple transmit coils, which could provide enough degrees of freedom to enable EP reconstruction using the relative phase of $B_{1}^{+}$  between the coil channels \cite{SodicksonGeneralizedLMT2013, serralles2019noninvasive}, which can be measured. This approach will be explored in future work. 

The current version of PIFON-EPT has a limitation when $B_z$ can not be assumed equal to zero. In this case, boundary artifacts appearing in the reconstructed EP cannot be eliminated. Previous work suggests that this limitation could be overcome by using multiple transmit-receive coils \cite{SodicksonGeneralizedLMT2013}. In this work, we used instead a birdcage coil, for which $B_z$ can be assumed negligible if the main field strength is lower or equal to 3T. However, we found that our network's expressive power was not enough to reconstruct both the EP and the $B_{1}^{+}$ in such a case. To address this, we made our network deeper and used more complex architectures (for example, we included Fourier mapping also in the EP Net) to accurately represent the EP and $B_{1}^{+}$, which ultimately increased the network's training time. This problem could be solved by designing compressed network architectures \cite{novikov2015tensorizing, liu2022tt} to replace the current fully-connected neural networks.
 
\section{Conclusion} \label{Conclusion}
We introduced PIFON-EPT, a new technique to estimate EP and magnetic transmit field distributions from noisy and/or incomplete MR measurements. We demonstrated our new approach using a series of numerical examples, showing that PIFON-EPT is accurate and robust even when its input is corrupted with a significant amount of noise. Since PIFON-EPT can efficiently de-noise MR measurements, it has the potential to improve other MR-based EPT methods that rely on magnetic transmit fields as inputs. In future work, we will investigate the performance of the proposed algorithms with realistic human head models and perform experimental validation.

\section*{Acknowledgments}
The authors are grateful to Alessandro Arduino for his valuable advice about the usage of EPTlib. 

\bibliographystyle{IEEEtran}
\bibliography{IEEEabrv,main}

\end{document}